\documentclass[lettersize,journal]{IEEEtran}
\usepackage{etoolbox}
\makeatletter
\@ifundefined{color@begingroup}%
{\let\color@begingroup\relax
\let\color@endgroup\relax}{}%
\def\fix@ieeecolor@hbox#1{%
\hbox{\color@begingroup#1\color@endgroup}}
\patchcmd\@makecaption{\hbox}{\fix@ieeecolor@hbox}{}{\FAILED}
\patchcmd\@makecaption{\hbox}{\fix@ieeecolor@hbox}{}{\FAILED}
\usepackage{amsmath,amsfonts}
\usepackage{algorithmic}
\usepackage{algorithm}
\usepackage{amsmath}
\usepackage{subfigure}  
\usepackage{array}
\usepackage[table,xcdraw]{xcolor}
\usepackage{multirow}
\usepackage[colorlinks, linkcolor=blue, citecolor=blue, urlcolor=blue, filecolor=blue]{hyperref}
\usepackage{mathrsfs}
\usepackage{textcomp}
\usepackage{stfloats}
\usepackage{url}
\usepackage{verbatim}
\usepackage{graphicx}
\usepackage{cite}
\usepackage{orcidlink}
\usepackage{cleveref}
\usepackage{subcaption}
\usepackage{float}
\begin{document}

\title{MID: A Comprehensive Shore-Based Dataset for Multi-Scale Dense Ship Occlusion and Interaction Scenarios}


\author{Yugang Chang\textsuperscript{\large\orcidlink{0009-0009-9247-8098}},
Hongyu Chen\textsuperscript{\large\orcidlink{0009-0004-2934-3135}}, 
Fei Wang\textsuperscript{\large\orcidlink{0009-0005-3061-0496}}, 
Chengcheng Chen\textsuperscript{\large\orcidlink{0000-0002-6014-5135}}, \IEEEmembership{Graduate Student Member, IEEE}, 
and Weiming Zeng\textsuperscript{\large\orcidlink{0000-0002-9035-8078}}, \IEEEmembership{Senior Member, IEEE}

\thanks{This work was supported by the National Natural Science Foundation of China (grant nos. 31870979), in part by the 2023 Graduate Top Innovative Talents Training Program at Shanghai Maritime University under Grant 2023YBR013. (Corresponding author: Weiming Zeng)}
\thanks{ Yugang Chang, Hongyu Chen, Fei Wang, Chengcheng Chen, Weiming Zeng are with the Laboratory of Digital Image and Intelligent Computation, Shanghai Maritime University, Shanghai 201306, China (e-mail: cygang\_post@163.com, hongychen676@gmail.com, shine\_wxf@163.com, shmtu\_ccc@163.com, zengwm86@163.com).}
\thanks{Manuscript received April 19, 2021; revised August 16, 2021.}}

\markboth{Journal of \LaTeX\ Class Files,~Vol.~14, No.~8, Dec~2024}%
{Shell \MakeLowercase{\textit{et al.}}: A Sample Article Using IEEEtran.cls for IEEE Journals}


\maketitle

\begin{abstract}
This paper introduces a maritime ship navigation behavior dataset, referred to as MID, which is annotated using Oriented Bounding Boxes (OBB) to address the challenges of ship target detection in complex real-world maritime environments. MID currently contains 5,673 images with a total of 135,884 finely annotated target instances, providing a robust foundation for both supervised and semi-supervised learning tasks. The dataset encompasses a variety of maritime scenarios, including ship encounters in diverse weather conditions, docking maneuvers, small target aggregation, and partial occlusions. These scenarios contribute to filling the gap in existing datasets (e.g., HRSID, SSDD, NWPU-10) by addressing the lack of data on complex situations such as occlusion and dense interactions, which are particularly relevant in the context of advanced technologies such as autonomous ships. All images in MID are sourced from high-definition video clips of real-world maritime navigation across 43 distinct navigable water areas. The dataset is further enriched through the inclusion of multiple weather and lighting conditions (e.g., rainy, cloudy, foggy days) and manually supplemented annotations, enhancing its diversity and ensuring that models trained on MID are better equipped to handle varying natural environments. This diversity also ensures its applicability to the real-world demands of busy ports and dense maritime regions. Using MID, we conducted a comprehensive evaluation of 10 detection algorithms, presenting a detailed analysis and comparison of their performance. This study includes: 1) an in-depth analysis of the dataset, 2) a showcase of detection results from various models, and 3) a comparison of baseline algorithms, with particular focus on their ability to detect dense occlusions. The results suggest that MID holds significant potential for advancing research in intelligent maritime traffic monitoring systems and autonomous vessel navigation, with important implications for future applications in maritime environment and safety situational awareness. Our dataset will be released at \href{https://github.com/VirtualNew/MID\_DataSet}{https://github.com/VirtualNew/MID\_DataSet}.

\end{abstract}


\section{Introduction}
\IEEEPARstart{I}{n} the context of the global development of smart ships and maritime port traffic management, the accuracy of target detection, tracking, and trajectory prediction for ships in complex scenarios is crucial for ensuring port safety, enhancing logistics efficiency, and reducing maritime accidents \cite{martelli2021outlook}. With the rapid growth of the global shipping industry, ports have increasingly become hubs of international trade \cite{peters2001developments}. Therefore, ensuring the safe navigation and efficient operation of ships entering and leaving ports in complex environments is especially important. Modern ports are typically equipped with numerous surveillance cameras to monitor the entry, exit, and activities of ships in real time \cite{yang2018internet}. However, the complexity of port environments characterized by a high number of ships, overlapping navigation paths, and diverse ship sizes and types poses significant challenges for ship detection and tracking \cite{xin2023multi}. In densely populated areas, smaller ships are particularly prone to being obscured or confused with other ships, further complicating detection and tracking tasks \cite{zhou2023sidelobe}. Addressing these issues is not only vital for the operational efficiency of ports but also plays a key role in reducing the incidence of maritime accidents.

At the same time, with the advancement of autonomous ship technology, intelligent navigation has become a research hotspot. The intelligent navigation of autonomous ships relies on three core technologies: environmental and safety situational awareness, route planning and collision avoidance decision-making, and ship motion control. Among these, environmental and safety situational awareness \cite{thombre2020sensors} is a prerequisite for autonomous navigation, particularly the reliable identification of the relative motion of external ship targets, which directly impacts navigational safety. In the early stages of autonomous ship development, navigational safety should be regarded as the primary objective of perception systems \cite{felski2020ocean}. However, the complexity of navigation environments and behaviors makes this task more challenging. When constructing datasets for ship target detection, it is essential to consider multiple influencing factors in complex navigation scenarios. First, the diversity of all-weather external environments, including different weather conditions (such as sunny, cloudy, rainy, and foggy days) and lighting conditions (daytime and nighttime), as well as background noise from sea spray and cloud shadows \cite{chen2024weather}. Second, the complex navigational behaviors between ships, such as crossing, overtaking, and meeting behaviors that occur in open waters and busy navigation areas, which may lead to occlusion and pixel overlap in images \cite{prasad2018object}. Finally, the variations in shape, size, and tonnage of different ships also affect the detection model, as differences in ship size result in significant variations in the pixel distribution of targets within images \cite{shi2013ship}. Therefore, effective perception and processing of these complex factors are crucial for achieving safe and efficient navigation in autonomous ship systems.

Whether in traditional port surveillance or in the intelligent navigation of autonomous ships, the core challenges of ship detection, tracking, and identification are similar. Both require accurate recognition of dense and occluded targets in complex environments. This necessitates not only advanced target detection technologies but also the development of robust datasets that can comprehensively cover the diverse navigation environments and ship behaviors.

Despite the widespread use of automatic identification systems (AIS) and radar for ship monitoring, they have significant limitations \cite{yang2019big}. AIS relies on ships actively transmitting signals, which means it cannot monitor ships that have not turned on their AIS or whose signals are disrupted \cite{barnum1986ship}. Although radar is effective in low visibility conditions, it struggles with target overlap and interference in dense environments, making it challenging to accurately distinguish complex interactions between ships \cite{robards2016conservation}. Additionally, radar has lower resolution and provides less detailed and intuitive information about ship targets, limiting its effectiveness for detecting smaller ships \cite{marino2015ship}. In contrast, optical images, with their high resolution, low cost, and rich visual information, have emerged as an ideal data source for ship target detection, particularly for real-time monitoring at medium to short distances \cite{zhang2017ship}. Compared to infrared thermal cameras, low-light night vision devices, and millimeter-wave radar, optical images offer more precise target detection in congested navigation areas and are relatively less affected by environmental and weather conditions \cite{wang2021research}. They can clearly capture the shape and position of ships, excelling in complex and dense scenarios. When combined with deep learning techniques, optical data can effectively address detection challenges posed by targets of various sizes, occlusion, and intricate interaction scenarios, providing robust support for the environmental perception of autonomous ships.

Moreover, existing maritime datasets often overlook the impact of marine weather conditions, such as rain, fog, overcast skies, and sunny days, as well as issues like image blurriness caused by weather \cite{hu2024dataset}. These conditions impose higher demands on ship detection and tracking, increasing the challenges posed by low visibility, varying lighting, and noise interference. Therefore, for small target detection and tracking in scenarios characterized by diverse sizes, dense occlusion, complex interactions, and varying marine weather conditions, optical data offers significant advantages and holds considerable research and application value.

Currently, visual tasks in dense maritime scenarios—such as overtaking, crossing, and meeting—are increasingly becoming an important research direction within the field of computer vision. Several existing datasets, such as HRSID (High-Resolution Ship Detection Dataset) \cite{wei2020hrsid}, SSDD (SAR Ship Detection Dataset) \cite{zhang2021sar}, and NWPU-10 \cite{su2019object}, \cite{su2020hq}, have provided foundational support for tasks related to ship detection, tracking, and prediction, advancing the technology for maritime traffic management and intelligent systems. However, despite the significant role these datasets play in enhancing the accuracy and applicability of computer vision tasks, specialized visual task datasets addressing the complexities of occlusion and dense interactions in maritime environments remain scarce. This is particularly critical in maritime applications involving advanced technologies like unmanned ships, where high-quality datasets specifically targeting complex ship interaction behaviors and dense environments are essential for achieving precise navigation and decision-making. Unfortunately, current datasets often overlook the unique demands of such complex scenarios. Therefore, filling this data gap will provide crucial support for unmanned ship technology and other maritime automation systems, further promoting the development of intelligent maritime traffic management.

Based on the aforementioned issues and the advantages of optical images, this paper presents the MID, constructed from visible light video images and demonstrating significant benefits. The dataset was meticulously annotated and processed over three months by four professionals in the port sector, ensuring high accuracy and detail. This high-quality annotation provides a reliable foundation for supervised learning, significantly enhancing model performance. Moreover, the detailed labeling system offers substantial potential for semi-supervised learning tasks, effectively guiding models to learn from a limited amount of labeled data and increasing the utilization of unlabeled data. Consequently, the MID is not only suitable for supervised learning tasks but also serves as an ideal experimental platform for semi-supervised learning research. It provides crucial data support for fields such as small target occlusion detection and ship tracking in maritime traffic management.

The significance of this research lies in the construction of the MID, which fills the gap in existing datasets regarding multi-scene dense occlusion targets, particularly providing crucial support for small target detection. This dataset encompasses dense ship scenarios and complex interaction behaviors—such as overtaking, crossing, and meeting—ensuring the precise recognition of small ships by detection algorithms in challenging environments, thereby offering robust technical assurance for the safe management of ports and maritime traffic. Additionally, the MID serves as an ideal testing platform for multi-object tracking tasks, with detailed annotations for scenes with severe occlusion or complex ship behaviors, significantly reducing the issue of target loss during tracking. Furthermore, the dataset provides essential support for the development of trajectory prediction models, covering a variety of ship interaction scenarios and possessing immense application value, especially in preventing ship collisions and managing complex behaviors. The construction of the MID offers critical support for optimizing detection, tracking, and trajectory prediction technologies, aiding in improving the performance of these visual techniques in complex port environments. By encompassing intricate ship interactions and dense occlusion scenarios, this dataset becomes an important resource for the study of unmanned ship technology, supporting the development and validation of autonomous navigation, obstacle avoidance, and trajectory prediction algorithms in complex marine environments, thus providing a vital data foundation for the practical application of unmanned maritime traffic systems.

\section{related work}

\subsection{Ship Detection Dataset}
Remote sensing optical datasets are typically obtained from satellite or aerial platforms, offering extensive coverage that can capture ship activities across large-scale areas such as oceans and ports. These datasets excel in monitoring widespread maritime traffic and capturing the distribution of ships along international shipping routes. The HRSID \cite{wei2020hrsid} focuses on ship detection tasks within complex backgrounds, providing precise ship target annotations that advance research in high-resolution ship detection within the remote sensing field. The SSDD \cite{zhang2021sar} is specifically designed for ship detection in synthetic aperture radar (SAR) imagery, with high-resolution images making it suitable for target detection and small target processing tasks. SSDD++ is an extended version of SSDD, demonstrating outstanding performance in detecting small targets within large-scale scenes and complex backgrounds. The HRSC2016 (High-Resolution Ship Collection Dataset 2016) \cite{liu2017high} offers detailed ship categories and accurate polygon annotations, making it suitable for tasks such as ship classification. It is a classic dataset in the field of remote sensing ship detection. ShipRSImageNet \cite{zhang2021shiprsimagenet} is a large-scale remote sensing image dataset that includes tens of thousands of images, specifically aimed at ship detection and classification tasks, achieving widespread success in multi-class ship detection in complex backgrounds. 

However, due to the relatively low spatial resolution of satellite imagery, especially at greater observation distances, detailed information about ship targets can be easily lost, leading to decreased detection accuracy. Additionally, remote sensing images are often affected by weather conditions such as cloud cover and haze, which can impact image quality and subsequently affect the extraction and recognition of ship targets. Furthermore, the infrequent capture of remote sensing images limits the ability to achieve real-time monitoring of ships at sea, restricting their application in dynamic scenarios.
\subsection{Optical Ship Detection Dataset}
Optical shore-based datasets are obtained from optical sensors fixed onshore, capturing high-definition images of ships in nearshore areas \cite{han2021fine}, \cite{shao2018seaships}. These datasets offer high spatial resolution, allowing for clear differentiation of detailed information such as ship shape, size, and heading. Particularly in densely populated areas with dynamic ship interactions, such as ports and shipping lanes, optical shore-based datasets can more accurately capture scenes involving multiple ships, providing rich information for subsequent visual tasks like ship classification and detection, tracking, and trajectory prediction. Additionally, because the data collection device is relatively fixed, continuous monitoring and high temporal resolution can be achieved, making it suitable for real-time monitoring needs. 

SeaShip \cite{shao2018seaships} is a typical optical shore-based ship detection dataset that includes images of various types of ships, primarily captured along coastlines and in port areas. This dataset provides multi-angle views of ships in complex backgrounds, making it especially suitable for target detection and tracking tasks in densely populated ship environments, thus offering important data support for the development and testing of related algorithms. However, there is a relative scarcity of datasets focusing on multi-scenario ship interaction behaviors and complex weather conditions specifically for optical shore-based ship detection in the current research field.
\subsection{Object Detection Algorithms}
Object detection algorithms have evolved from traditional methods to modern deep learning-driven approaches. Early target detection relied on handcrafted features, such as Histogram of Oriented Gradients (HOG) combined with SVM classifiers and methods like Deformable Parts Model (DPM) \cite{felzenszwalb2008discriminatively}. These algorithms extracted geometric features or decomposed target structures, achieving good detection performance under specific conditions. However, their generalization ability was limited in complex backgrounds or across different scales, and they often had high computational costs.

With the rapid advancement of deep learning, convolutional neural network based (CNN) object detection algorithms have become mainstream. The R-CNN series of methods (including R-CNN \cite{girshick2014rich}, Fast R-CNN \cite{girshick2015fast}, and Faster R-CNN \cite{ren2016faster}) introduced the Region Proposal Network (RPN), significantly improving both the accuracy and speed of object detection. R-CNN generates candidate regions first and then classifies and regresses them using a CNN to achieve object detection. Single-stage methods like You Only Look Once (YOLO) \cite{redmon2016you} and Single Shot Multibox Detector (SSD) \cite{liu2016ssd} further enhance real-time performance. YOLO treats object detection as a single regression problem, enabling end-to-end detection, while SSD combines multi-scale feature maps to improve detection of objects at different sizes. RetinaNet \cite{ross2017focal} introduced Focal Loss to address the class imbalance problem, and DEtection TRansformer (DETR) \cite{carion2020end} was the first to apply the Transformer architecture to object detection tasks, eliminating traditional region proposal and non-maximum suppression (NMS) steps. DETR uses a global attention mechanism to directly predict the classes and positions of objects. To address the slow training issue of DETR, Deformable DETR introduced a deformable convolutional network to accelerate convergence and improve detection efficiency. EfficientDet \cite{tan2020efficientdet} achieved a balance between high performance and low computational overhead through efficient network architecture design. These modern algorithms are widely applied in scenarios such as autonomous driving, video surveillance, and drone detection, continuously driving the development of object detection technology.

Based on the aforementioned related work, these advanced object detection algorithms have made significant progress in optimizing performance. However, high-quality training data is also a key factor in enhancing detection effectiveness. When constructing datasets, it is crucial not only to ensure data quality and detailed annotations but also to encompass complex weather conditions and diverse ship interaction scenarios that are essential for specific perception tasks. This is particularly important in the context of environmental perception for autonomous unmanned ships and maritime regulatory systems, where data from these complex scenes can effectively enhance the robustness and generalization ability of algorithms in real-world applications.

The core of our work is to establish a large dataset similar to existing visual tasks, but unlike current datasets, our target domain focuses on the perception of maritime ship navigation behavior, particularly emphasizing complex scenarios of ship interactions. This new dataset will provide more comprehensive support for research and applications in related fields, advancing the development of unmanned ship technology and maritime regulatory systems.

\section{Data Collection}
\begin{figure}
\small 
	\centering
	\includegraphics[width=\linewidth]{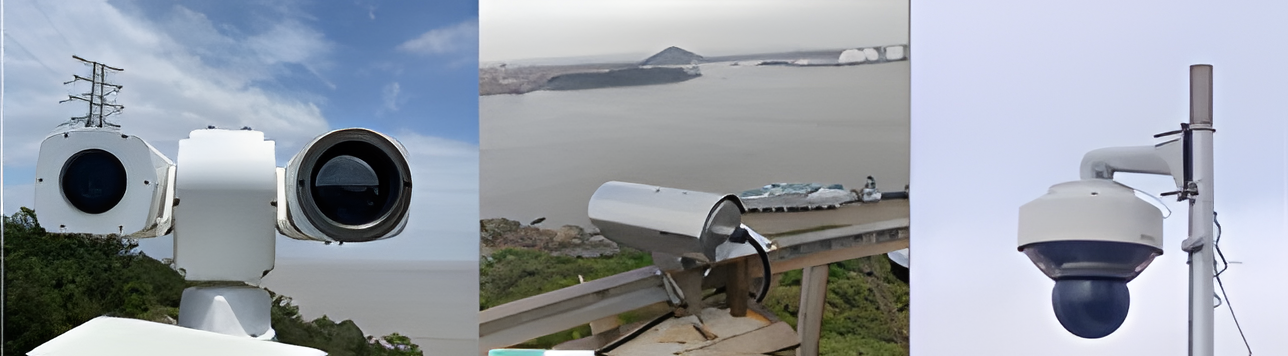}
	\caption{Video capture device.}
	\label{fig:fig1}
	\medskip
\end{figure}

We constructed the MID based on images captured by video monitoring systems deployed in real environments. Fig. \ref{fig:fig1} shows the video capture device used, with specific parameters listed in Table \ref{tab:table1}. Three types of high-definition cameras were installed near open water areas of the port and in densely navigated narrow channels, covering a total area of 41 square kilometers. These cameras provide high-quality monitoring videos (1920×1080 pixels) from which the images for our proposed dataset are extracted. Specifically, one high-definition electro-optical device was installed on a platform at a height of 100 meters above the open water area, capturing images of ships entering and leaving the port, including a visible light zoom camera and a controllable rotating gimbal. In the densely navigated narrow channel area, one variable focal length camera and one panoramic fisheye camera were installed on a building at a height of 118 meters to capture images of ships engaged in crossing, overtaking, and encountering maneuvers.

Considering the relatively stable ship traffic in the open waters of the port and the densely navigated narrow channel areas throughout the year, we randomly selected March for continuous observation and data collection from both areas. We chose 4 days between March 1st and 15th, covering 16 scene video segments (with a total observation time of nearly 14 hours), and 6 days between March 16th and 27th, covering 27 scene video segments (with an observation time of nearly 17.5 hours). The selected scenes began as early as 07:49 Beijing time and ended as late as 18:08, effectively covering all time periods from morning to evening. In total, we obtained 43 video segments featuring a rich variety of real-world scenes.

\begin{table*}[htbp]
    \caption{Capture device parameter comparison table}
    \label{tab:table1}
    \centering
    \footnotesize
    \setlength{\tabcolsep}{5pt}
    \begin{tabular}{p{3.5cm}p{6cm}|p{6.5cm}}
    \hline
         Device Name & Device Parameters & Collection area and content \\
    \hline
         \multirow{9}*{Electro-Optical Device} &  \textbf{Visible Light Probe Specifications}&\\
         & 1.\textbf{Focal Length}: 10-860mm & \\
         & 2. \textbf{Resolution}: 1920 $\times$ 1080P& \\
         & 3. \textbf{Optical Zoom}: 86$\times$&\textbf{Open Water Area of the Port}\\
         & 4. \textbf{Detection Range}: Greater than 10km& \textbf{Scenes Included}:\\
    \cline{2-2}
         & \textbf{Gimbal Specifications}&1.	Ships approaching and departing from docks.\\
         & 1. \textbf{Rotation}: Continuous 360°& 2.	Overtaking, crossing, and encountering maneuvers\\
         & 2. \textbf{Horizontal Speed}: 0.01-45°/s& of ships at sea.\\
         & 3. \textbf{Positioning Accuracy}: 0.01° &\\
    \cline{1-3}
       \multirow{4}*{Variable Focal Length Camera} & \textbf{Lens Specifications}:&\\
       & 1. \textbf{Focal Length}: 10-770mm & \\
       & 2. \textbf{Optical Zoom}: 77 $\times$& \\
       & 3. \textbf{Maximum Resolution Supported}: 3840 $\times$ 2160P &\textbf{Densely Navigated Narrow Channel Area}\\
    \cline{1-2}
    \multirow{7}*{Panoramic Fisheye Camera} & \textbf{Sensor and Lens Specifications} & \textbf{Scenes Included}:\\
    & 1. \textbf{Sensor}: Composed of 4 high-definition Progressive Scan CMOS sensors & 1. Multi-scale images of different ships and ship types.\quad\quad 2. Ship occlusions in dense areas.\\
    & 2. \textbf{Maximum Resolution}: 1920 $\times$ 1080P &3. Panoramic images of both large and small ship targets.\\
    & 3. \textbf{Focal Length}: 7.1$\sim$320mm &\\
    & 4. \textbf{Panoramic Monitoring}: & \\
    & \quad a) Horizontal Field of View: 180°&\\
    & \quad b) Vertical Field of View: 85°&\\
    \hline
    \end{tabular}
\end{table*}

\begin{figure}
\small 
	\centering
	\includegraphics[width=\linewidth]{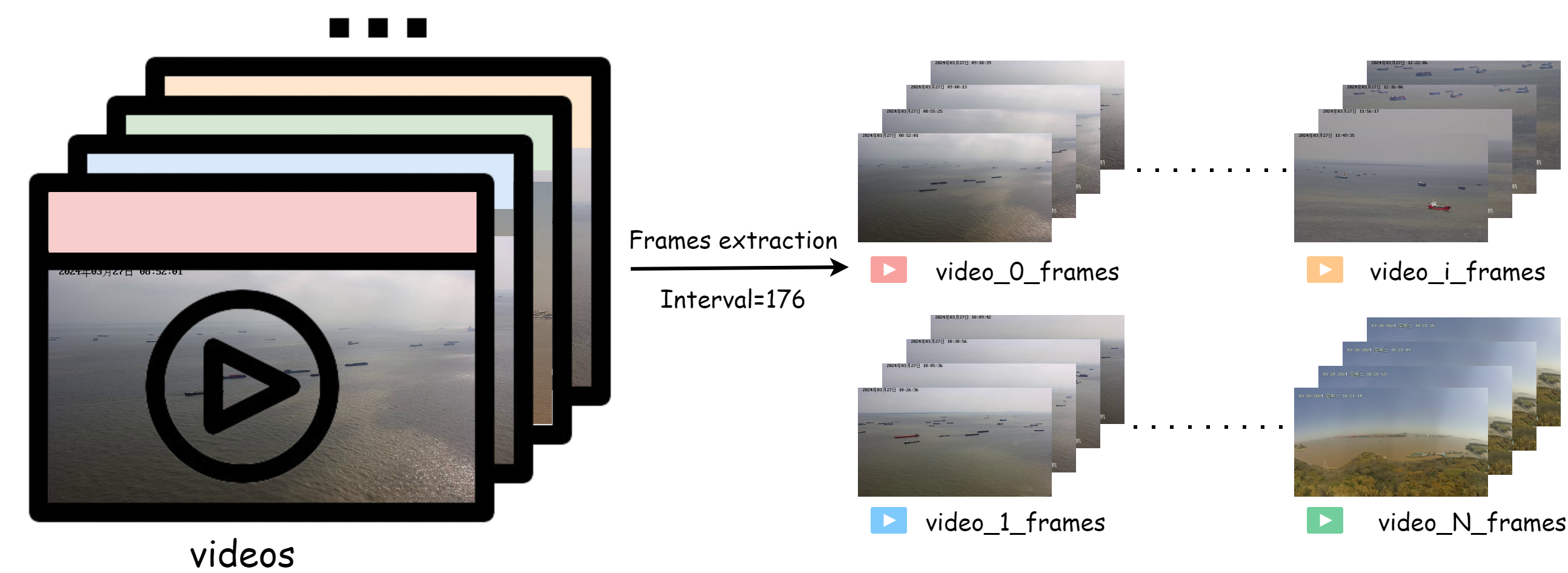}
	\caption{Flow chart for extracting video frames.}
	\label{fig:fig2}
	\medskip
\end{figure}

\begin{figure}
\small 
	\centering
	\includegraphics[width=0.9\linewidth]{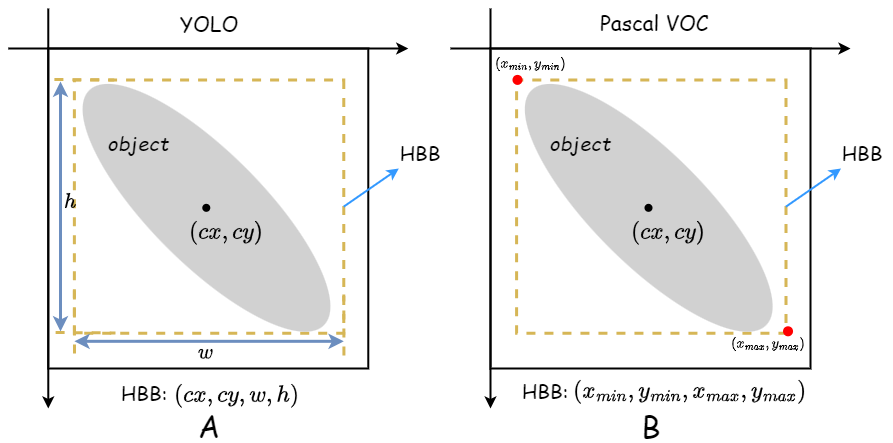}
	\caption{HBB annotation method.}
	\label{fig:fig3}
	\medskip
\end{figure}

\begin{figure}
\small 
	\centering
	\includegraphics[width=0.5\linewidth]{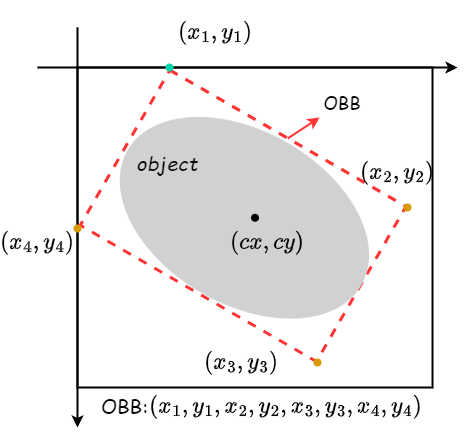}
	\caption{OBB annotation method.}
	\label{fig:fig4}
	\medskip
\end{figure}

\subsection{Visible Light Video Image Production}
The dataset we propose is derived from the video sequences recorded by front-end cameras, from which visible light image frames are extracted. The panoramic fisheye camera is used to capture a wide range of video data. The visible light zoom camera and the pan-tilt-zoom (PTZ) camera not only capture high-definition video images in a single direction but can also rotate at any angle to obtain video from different perspectives, while allowing for zoom adjustments to adapt to various scales.

We divide each recorded video into different frames, using an inter-frame extraction mechanism that extracts one frame every 176 frames, as shown in Fig. \ref{fig:fig2}.

This provides a rich set of ship navigation images in complex real-world scenarios, facilitating the creation of a ship recognition dataset. We extracted ship navigation videos from 43 scene segments in open port waters and densely trafficked narrow channels. Each video has a recording duration of approximately 35 minutes, and we extracted one frame every 6 seconds, resulting in 350 images per video. In total, we obtained 15,050 original high-definition images depicting real navigation scenarios of ships.

The extracted frame images were accurately annotated by professionals with experience in ship navigation management and labeling, ensuring the quality and consistency of the annotations. Using Horizontal Bounding Boxes (HBB) can lead to inaccuracies in enclosing complex-shaped objects, resulting in significant area errors that affect the precision and effectiveness of object detection, as well as introducing redundant information, as shown in Fig. \ref{fig:fig3}. Therefore, we employed the Oriented Bounding Box (OBB) annotation method illustrated in Fig. \ref{fig:fig4}, with the output fields formatted as ($x_1, y_1, x_2, y_2, x_3, y_3, x_4, y_4$).

\begin{figure}
\small 
	\centering
	\includegraphics[width=0.9\linewidth]{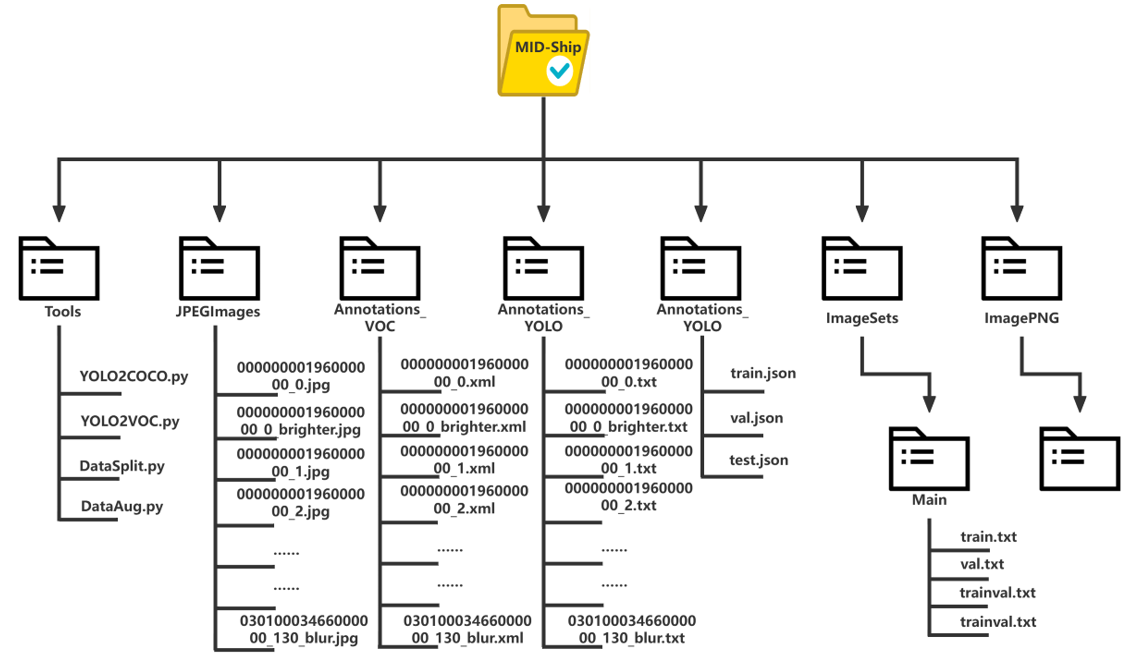}
	\caption{Overall file structure of MID.}
	\label{fig:fig5}
	\medskip
\end{figure}

In addition, we systematically organized the images and their annotation files in MID, dividing them into training, validation, and testing sets. We also included conversion scripts for various annotation formats in the toolkit to support the needs of different detection networks. The specific file structure of MID is detailed in Fig. \ref{fig:fig5}.

\subsection{Dataset Diversity}
\begin{figure*}
\small 
	\centering
	\includegraphics[width=\linewidth]{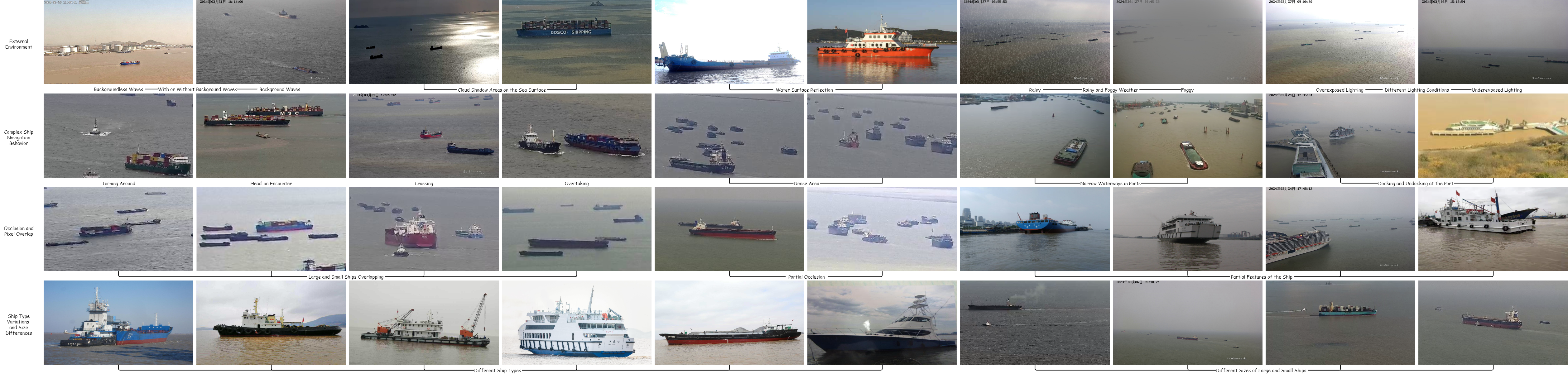}
	\caption{Diversity of MID.}
	\label{fig:fig6}
	\medskip
\end{figure*}

A good dataset should adequately reflect the complexity of the marine environment to ensure diversity while considering the unique characteristics of real-world scenes. This approach facilitates stable and consistent testing results from detection models. In the process of creating the ship recognition dataset, we collected real, complex navigational scenes and will take the following steps to ensure the diversity of the dataset. For details, please see Fig. \ref{fig:fig6}.
\subsubsection{Selection of External Environments}
In most detection tasks, especially in facial recognition, detection accuracy is rarely affected by background changes because the facial area is contained within a regular rectangle, making it easy to separate from the background. However, unlike faces, the process of marking bounding boxes for ships navigating on water surfaces often encounters significant background noise due to factors such as wave splashes, shadows from clouds on the sea surface, and reflections on the water. These background elements can be misidentified as features of the ship, impacting the final detection accuracy. To mitigate the effects of a single background, we collected ship navigation image data under various conditions, including clear days at different times, cloudy occlusions, wave splashes, and rainy or foggy conditions with varying lighting.
\subsubsection{Extraction of Complex Navigation Behaviors Between Ships}
When ships navigate at sea, scenarios such as turning, meeting, crossing, and overtaking inevitably occur. These situations reflect the complex navigation behaviors of ships, as they must adhere to international maritime collision avoidance regulations. Therefore, we selected high-definition images of multiple ships entering and leaving the port in open waters, as well as images depicting ship navigation dynamics under high traffic density in narrow waterways. This approach ensures that the ship identification dataset covers a diverse range of unique features across different real-world scenarios.
\subsubsection{Partial Occlusion and Pixel Overlap}
Partial occlusion and pixel overlap are common issues in maritime imaging. As most ships are moving and the pan-tilt cameras rotate, we often observe only parts of the ships entering and exiting the camera's field of view. In reality, these partially occluded ships are still objects that need to be detected. Therefore, we annotate both the visible parts of the hull and the obscured sections at different visibility ratios. Additionally, due to the installation angle of the camera, ships navigating close to each other may experience a certain degree of pixel overlap, where smaller ships can be obscured by larger ones. Ignoring this overlap would be unreasonable. Thus, we collected as much occluded data as possible to ensure that the subsequent training models can effectively handle such occlusions.
\subsubsection{The diversity of shapes and tonnages of ships navigating on the sea}
The differences in pixel representation of targets in the same image are caused by the various shapes and sizes of ships. In such scenes, some ships may not always be positioned well within the optimal pixel area. We collected data on large ships along with multiple smaller ships in their vicinity, as well as different types of ships such as cruise ships, cargo ships, and container ships. This expansion enhances the diversity of the ship identification dataset, allowing for effective detection across a range of scales.

\begin{figure}
\small 
	\centering
	\includegraphics[width=\linewidth]{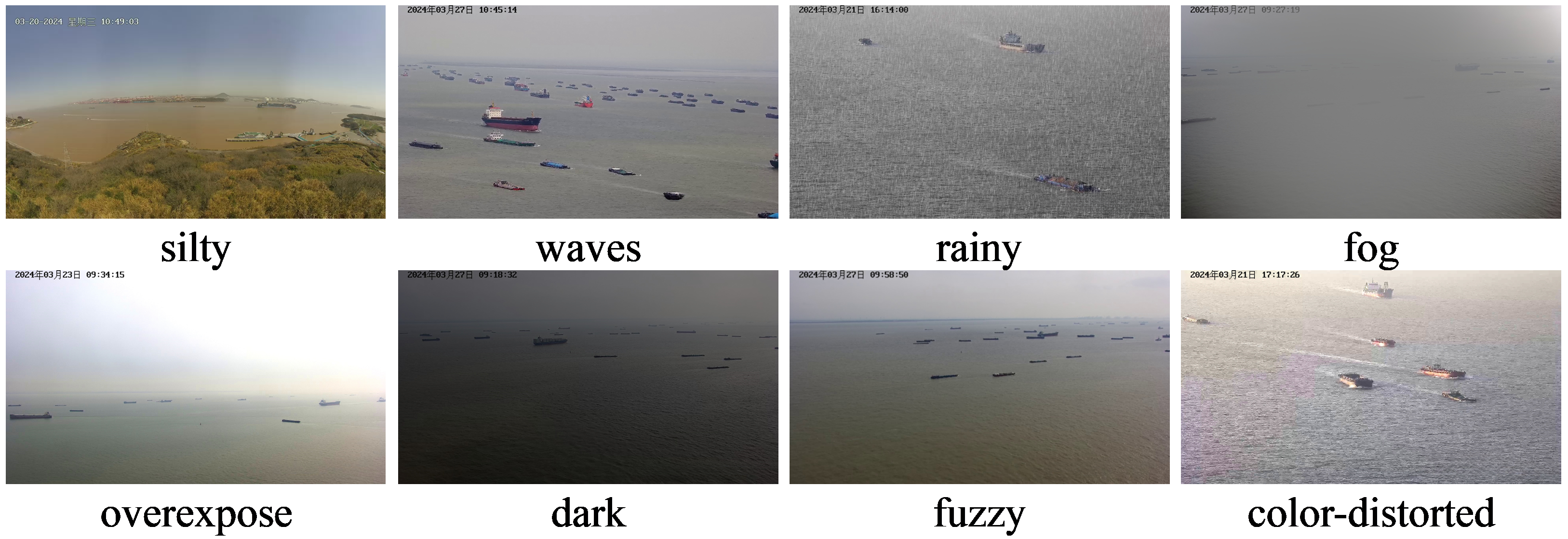}
	\caption{Examples images of different weather variations in MID. The 8 sub-images correspond to 8 different weather scenes.}
	\label{fig:fig7}
	\medskip
\end{figure}
\begin{figure}
\small 
	\centering
	\includegraphics[width=\linewidth]{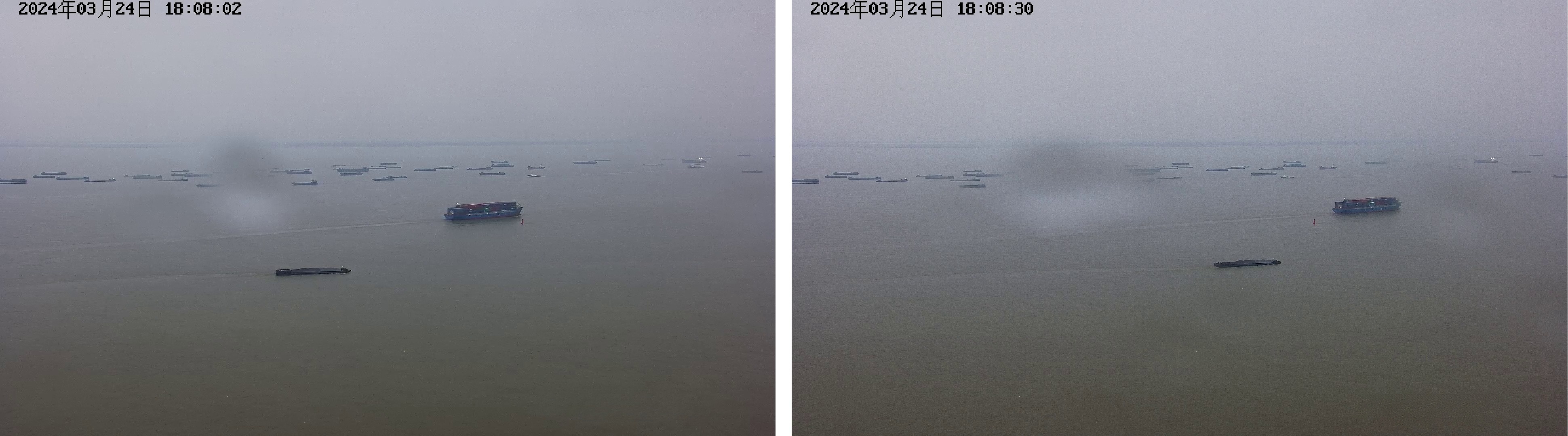}
	\caption{Example images of water droplets appearing on cameras in MID.}
	\label{fig:fig8}
	\medskip
\end{figure}

\begin{figure}
\small 
	\centering
	\includegraphics[width=0.9\linewidth]{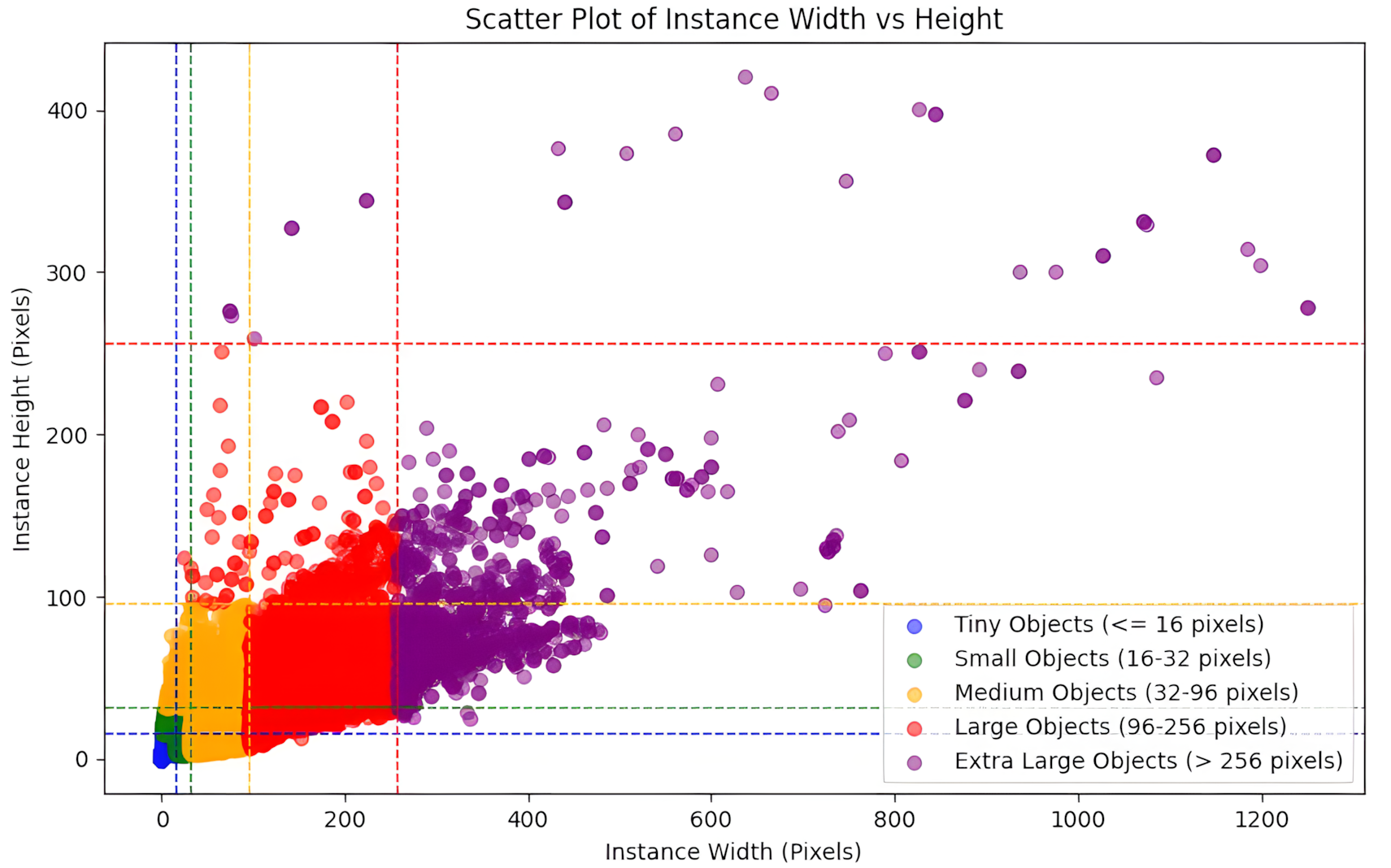}
	\caption{Scatter plot of instance width vs height in MID.}
	\label{fig:fig9}
	\medskip
\end{figure}
\begin{figure}
\small 
	\centering
	\includegraphics[width=0.8\linewidth]{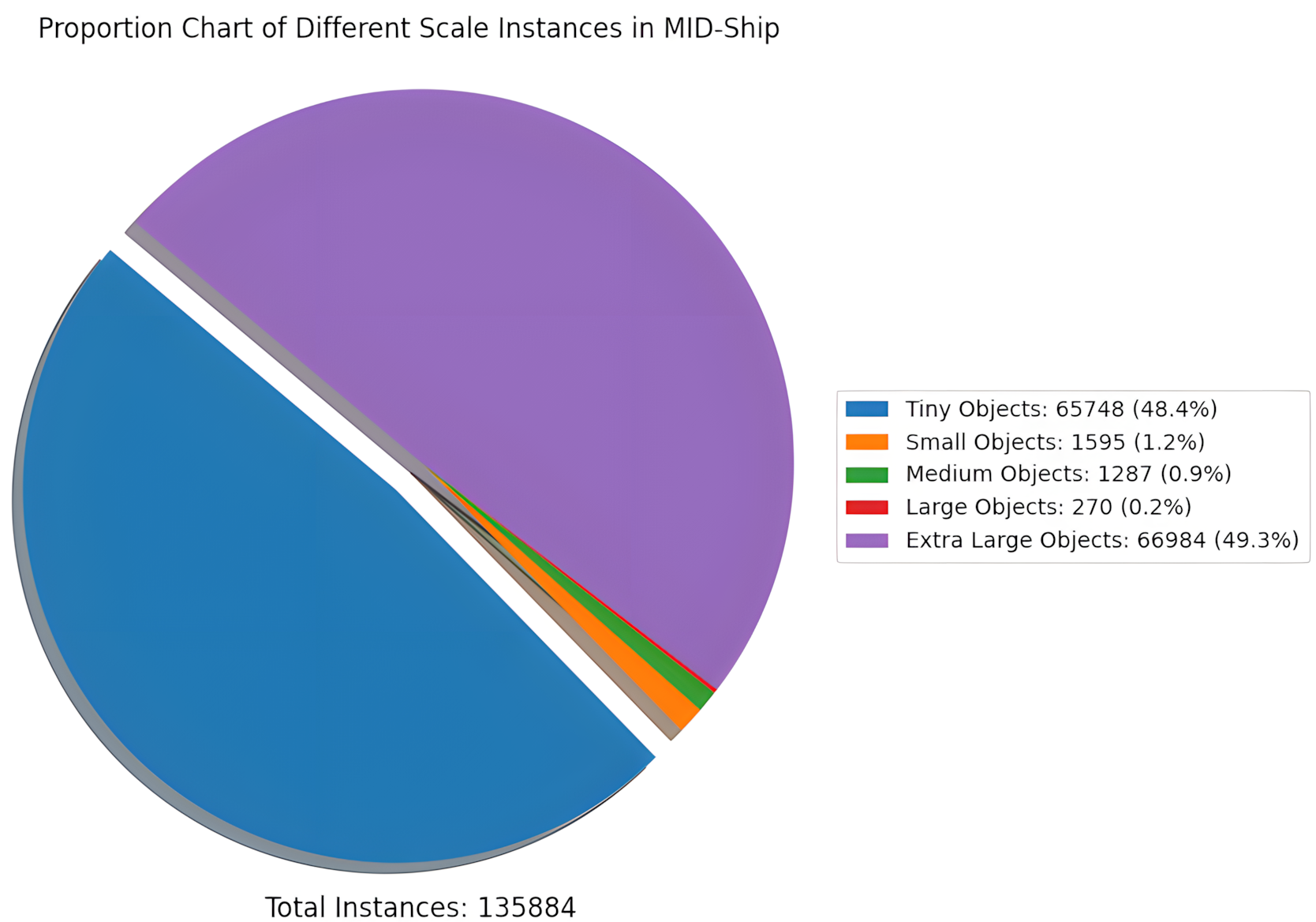}
	\caption{Proportion chart of different scale instances in MID.}
	\label{fig:fig10}
	\medskip
\end{figure}
\begin{figure}
\small 
	\centering
	\includegraphics[width=0.95\linewidth]{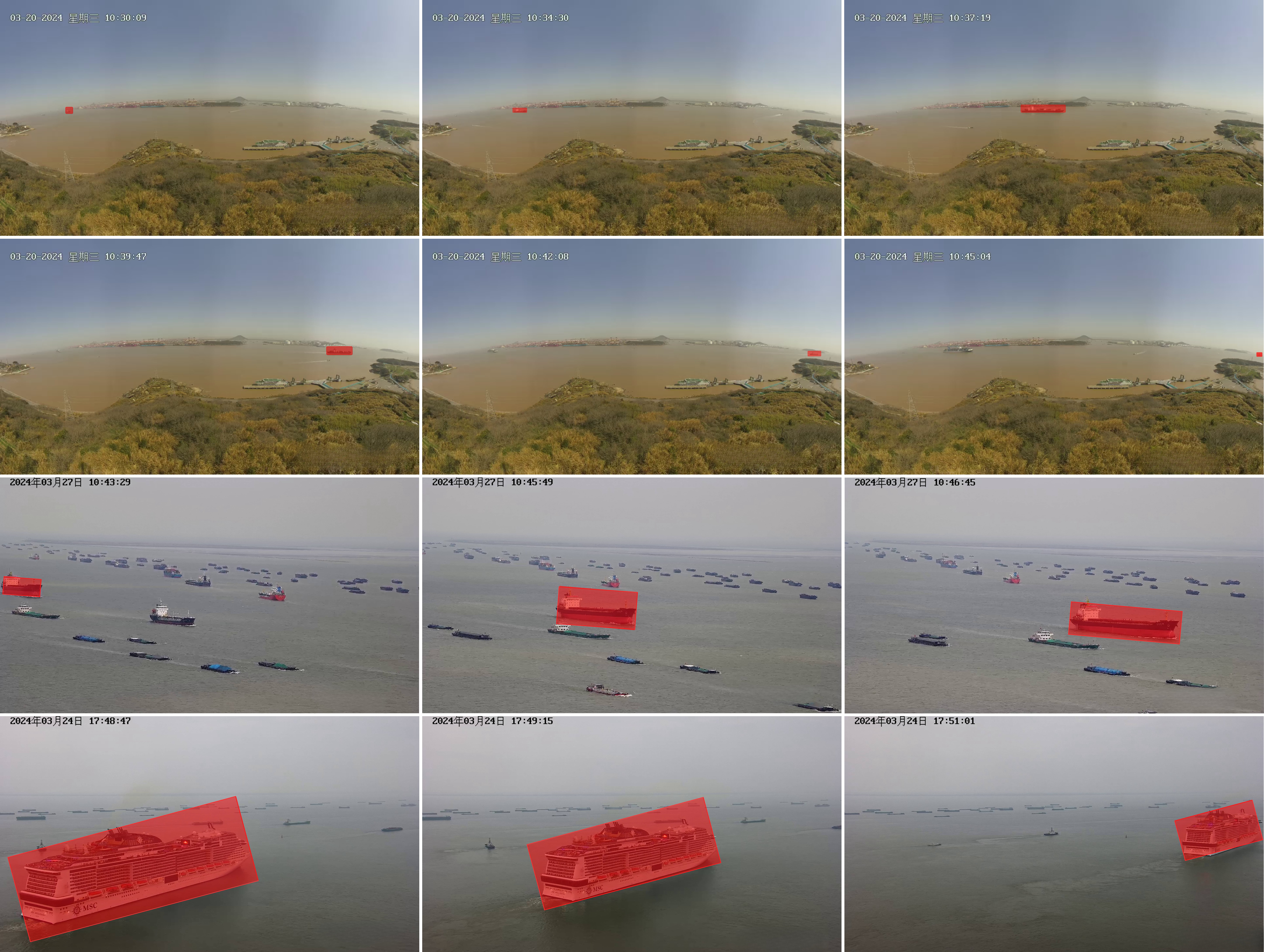}
	\caption{The example images in the first to second rows show six different sub viewpoints of the same ship in MID in a panoramic view. The third and fourth rows are example images of scale changes in two scenes, with close-up views, respectively. The two rows of images show scale changes in two different scenes.}
	\label{fig:fig11}
	\medskip
\end{figure}
\begin{figure}
\small 
	\centering
	\includegraphics[width=0.7\linewidth]{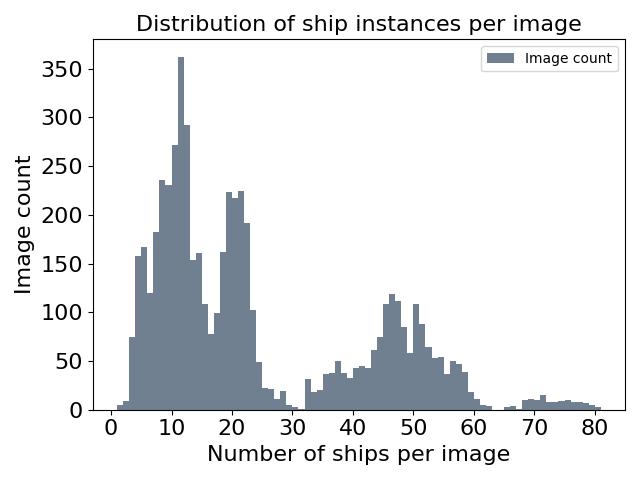}
	\caption{Distribution of ship instances of each image in MID.}
	\label{fig:fig12}
	\medskip
\end{figure}

\section{DESIGN AND ANALYSIS OF MID}
To highlight the advantages of MID, we conduct an in-depth analysis from six key aspects: weather conditions, target scale, target aspect ratio, background interference, target occlusion, and ship collision avoidance. These factors significantly impact the performance of object detection, reflecting the diversity and application potential of the dataset.
\subsection{Weather Variation}
Due to the diversity of weather conditions, which provides richer visual effects, the image dataset becomes more representative and can better reflect various real-world scenes. Therefore, we analyzed the types of weather conditions in MID and categorized and analyzed the data for each type of scene. We classified weather conditions into eight categories: silty, waves, rainy, fog, overexposed, dark, fuzzy, and color-distorted, as shown in Fig. \ref{fig:fig7}.

We further analyzed the shooting equipment and observed that, in a small portion of images, water droplets appeared on the camera lens due to adverse weather conditions, as illustrated in Fig. \ref{fig:fig8}. This detail adds valuable diversity to the dataset and increases its complexity, as varying environmental conditions can greatly influence camera performance. Incorporating these factors into the model-building process enhances the model’s ability to adapt to complex real-world scenarios, ultimately boosting its effectiveness in practical applications.

\subsection{Scale Variation}
Scale variation is an important characteristic, reflecting how ships of different sizes appear in images. Due to varying shooting conditions and perspectives, the same vessel may present different scales across images. This variation can stem from factors such as shooting distance, lens focal length, and environmental conditions. To explore scale variation in MID, we first examined the overall scale distribution in the dataset, as shown in Table \ref{tab:table2}. Fig. \ref{fig:fig9} and \ref{fig:fig10} illustrate the scale of labeled targets more vividly through scatter plots and proportion charts. We then broadly categorized the shooting scenes into two types: panoramic and close-up. Fig. \ref{fig:fig11} describes the scale variation in both close-up and distant views. 

Fig. \ref{fig:fig12} shows the distribution of ship instances per image in MID, where most images contain between 5-25 instances, and a few contain 35-55 instances. On average, each image includes 23.953 ship instances, with a maximum of 81 and a minimum of 1 instance per image.

\begin{figure}
\small 
	\centering
	\includegraphics[width=0.7\linewidth]{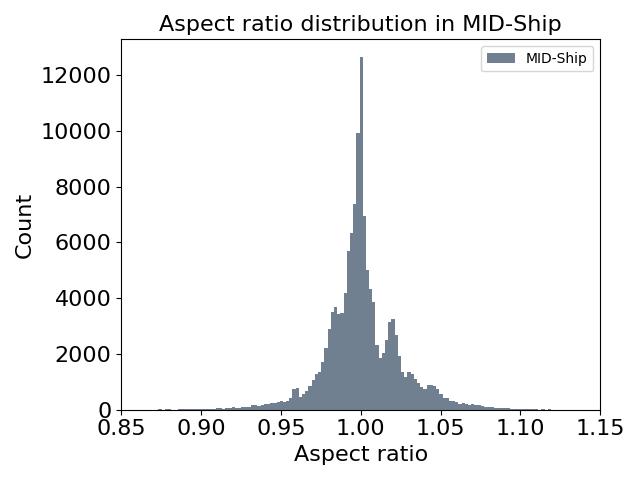}
	\caption{AR distribution in MID.}
	\label{fig:fig13}
	\medskip
\end{figure}

\begin{table}[htbp]
    \caption{Scale variation statistics in MID.}
    \label{tab:table2}
    \centering
    \footnotesize
    \setlength{\tabcolsep}{18pt}
    \begin{tabular}{ccc}
    \hline
      Instances    & Illustration & Number \\
    \hline
         Tiny Instances & $\leq$ 16 pixels & 65748\\
         Small Instances & 16-32 pixels & 1595\\
         Medium Instances & 32-96 pixels & 1287\\
         Large Instances & 96-256 pixels & 270\\
         Extra Large Instances & $>$ 256 pixels & 66984 \\
         \cline{1-3}
         Total Instances &  - & 135884 \\
    \hline
    \end{tabular}
\end{table}

\begin{figure}
\small 
	\centering
	\includegraphics[width=\linewidth]{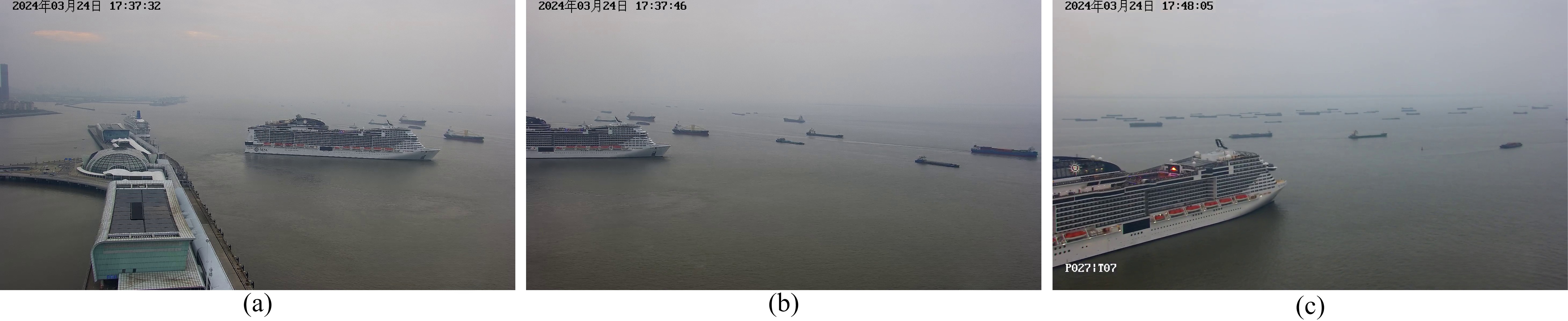}
	\caption{Three different viewpoints of the same fishing boat: looking to the left (a), looking to the center (b), and looking to the right (c).}
	\label{fig:fig14}
	\medskip
\end{figure}

\subsection{AR of Instance Variations}
The aspect ratio (AR) of ship objects is an important feature, typically ranging from 3 to 6, and is one of the key characteristics for ship classification and identification tasks. For deep learning-based methods, AR serves as a crucial parameter when designing anchor-based models, such as Faster R-CNN \cite{ren2016faster}. Fig. \ref{fig:fig13} illustrates the distribution of ship aspect ratios in MID.
\subsection{Viewpoint Variation}
In fact, a fixed-position camera can capture images of the same ship from different angles over the same body of water by rotating the high-definition camera. In this paper, we primarily use three perspectives: left, center, and right. Fig. \ref{fig:fig14} (a)-(c) illustrate the same ship from different sub-perspectives.

\begin{figure}
\small 
	\centering
	\includegraphics[width=\linewidth]{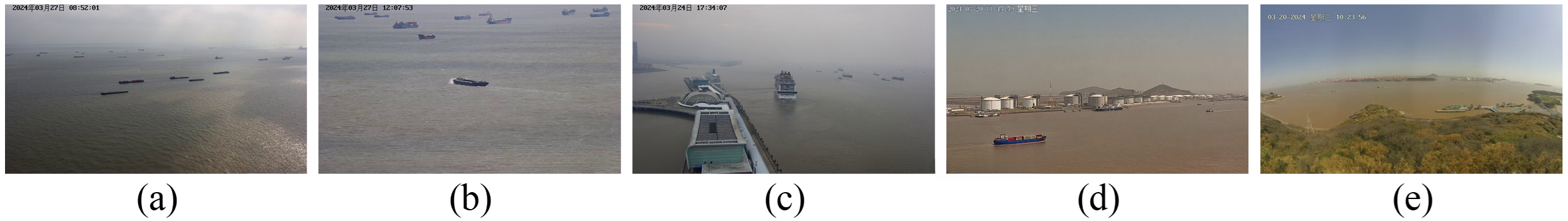}
	\caption{Five main different backgrounds in MID.}
	\label{fig:fig15}
	\medskip
\end{figure}

\begin{figure}
\small 
	\centering
	\includegraphics[width=0.85\linewidth]{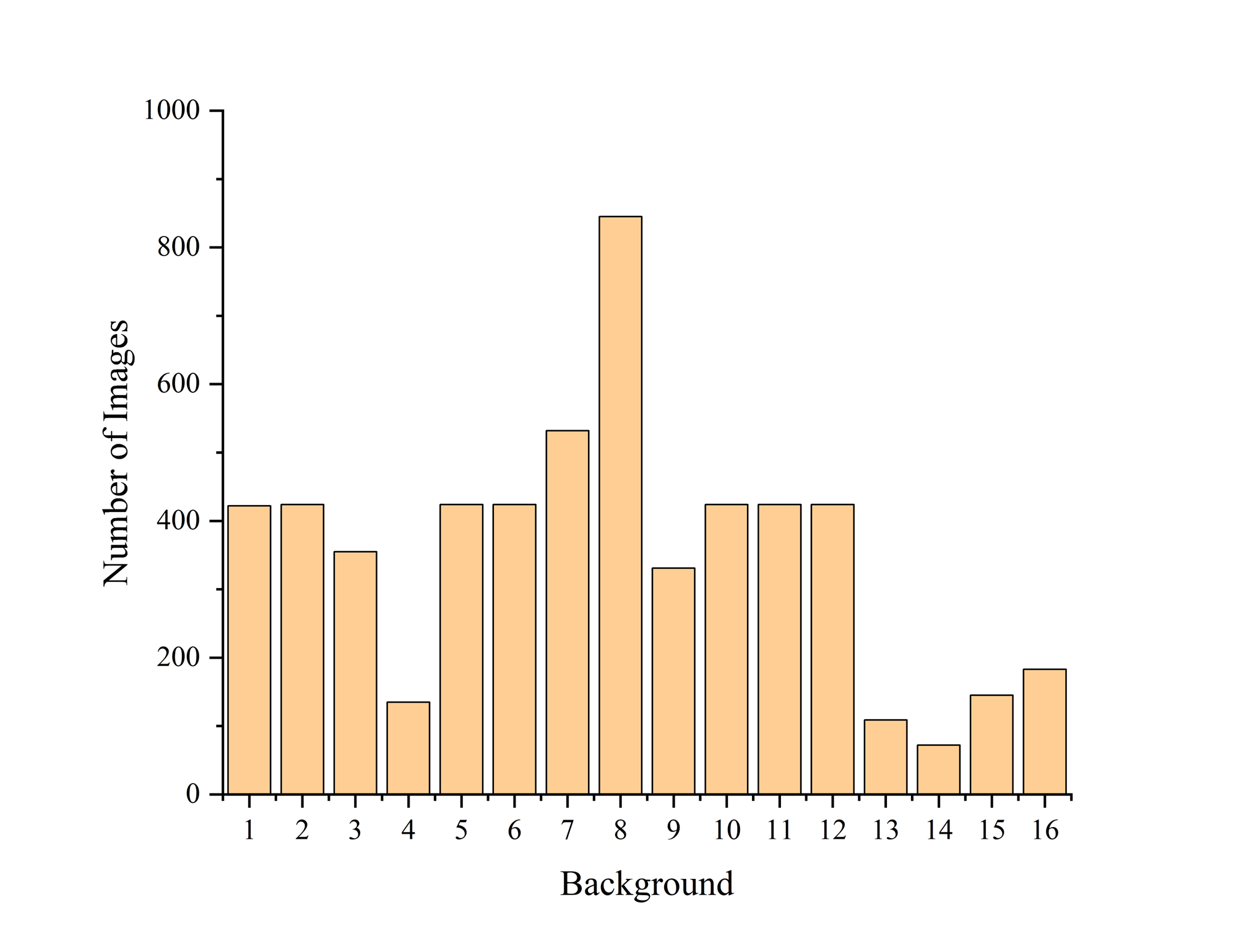}
	\caption{Sixteen different backgrounds in MID based on the number of the videos.}
	\label{fig:fig16}
	\medskip
\end{figure}
\begin{figure}
\small 
	\centering
	\includegraphics[width=\linewidth]{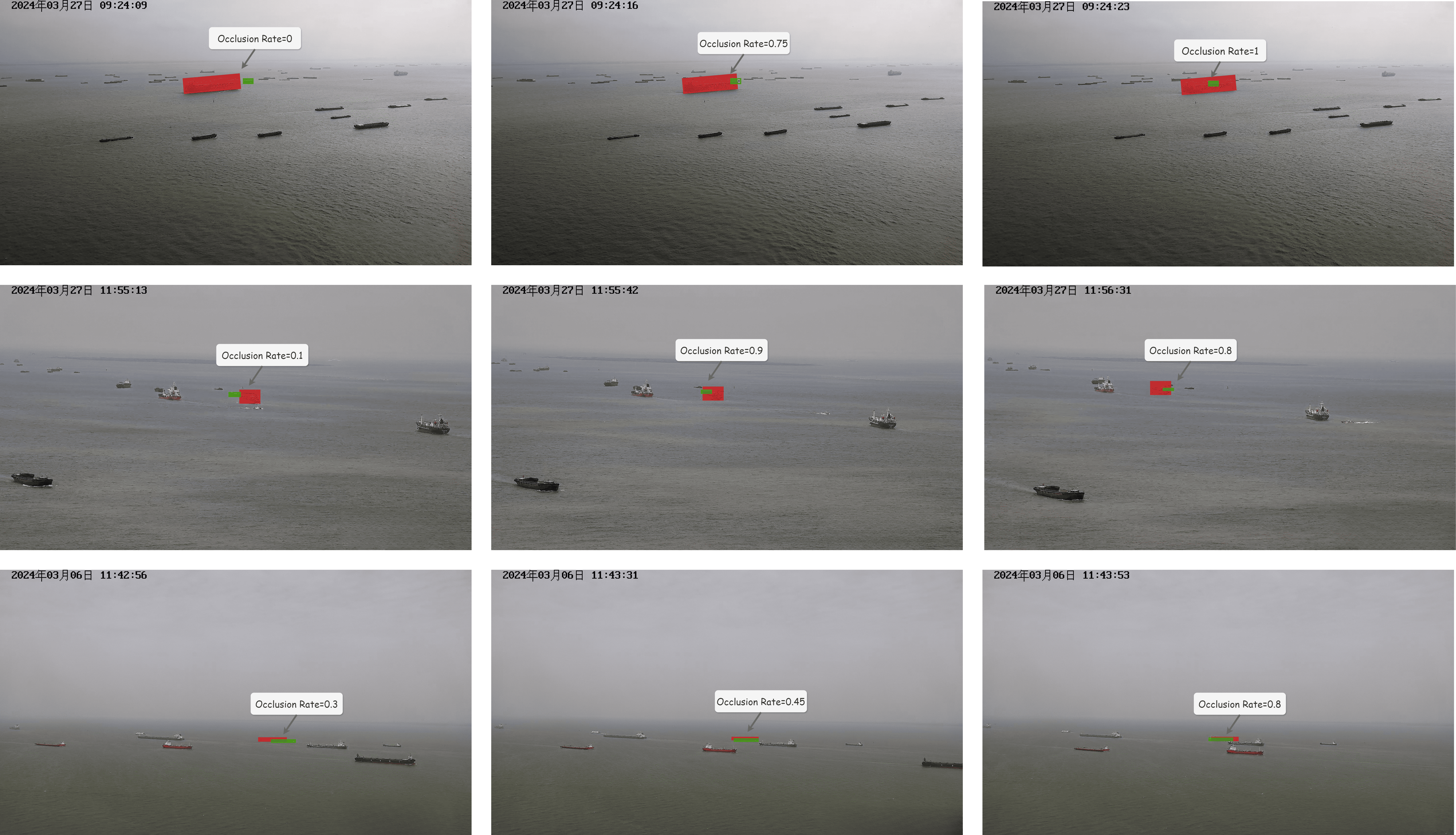}
	\caption{Examples images of different occlusion degrees. The first to third rows correspond to the occlusion conditions in three different scenarios.}
	\label{fig:fig17}
	\medskip
\end{figure}

\subsection{Background Variation}
The background diversity can be investigated based on the cameras deployed at different locations. To this end, we conducted an analysis of MID based on shooting angles and scenes, categorizing the backgrounds into five main types, as shown in Fig. \ref{fig:fig15}. Additionally, based on the number of video files captured, we classified the backgrounds into 16 categories and counted the number of images under each background, as shown in Fig. \ref{fig:fig16}.

\subsection{Occlusion Variation}
To investigate the types of occlusion in MID, we grouped all images with occlusion across different scenes and analyzed the degree of occlusion, as shown in Table \ref{tab:table3}. Our findings reveal that 16\% of the dataset contains varying levels of occlusion, with some images showing over 90\% occlusion or even complete occlusion. This makes MID a particularly challenging dataset for handling occlusions. Fig. \ref{fig:fig17} provides a visual representation of the distribution of ships under different levels of occlusion.

\begin{table}[htbp]
    \caption{Occlusion instance statistics}
    \label{tab:table3}
    \centering
    \footnotesize
    \setlength{\tabcolsep}{13pt}
    \begin{tabular}{ccc}
    \hline
      Instances    & Illustration & Number \\
    \hline
         Unobstructed Instances & $\leq$ 10\% & 114090\\
         Slightly Obstructed Instances & 10\%-20\% & 9287\\
         Partially Obstructed Instances & 20\%-50\% & 10036\\
         Severely Obstructed Instances & 50\%-90\% & 1634\\
         Completely Obstructed Instances & $\geq$ 90\% & 837 \\
         \cline{1-3}
         Total Instances &  - & 135884 \\
    \hline
    \end{tabular}
\end{table}

\begin{table}[htbp]
    \caption{Experimental Parameter Settings}
    \label{tab:table4}
    \centering
    \footnotesize
    \setlength{\tabcolsep}{8pt}
    \begin{tabular}{cccccc}
    \hline
        Epochs& Optimizer & Batch & LR0 &Weight\_decay & Imgsz\\
    \hline
         100 & AdamW & 16 & 0.002 & 5e-4 &640\\
    \hline
    \end{tabular}
\end{table}

\begin{table*}[htbp]
    \caption{Experimental Results}
    \label{tab:table5}
    \centering
    \footnotesize
    \setlength{\tabcolsep}{17.5pt}
    \begin{tabular}{ccccccc}
    \hline
         Method & P & R &  $\mathrm{mAP}_{.50}$(\%) & $\mathrm{mAP}_{.50:.95}$(\%) & Param(m) & Flops(G)\\
    \hline
         Yolov6n-obb\_head & 0.887 &0.688 & 79.5& 53.4&4.23&11.8\\
         Yolov6s-obb\_head & 0.916 &0.712 & 83.3& 59.1&16.37&44.3\\
         Yolov8n-obb & 0.912 &0.708 & 82.9 &58.8&3.01&8.3\\
         Yolov8s-obb & 0.935 &0.72 &85.2 &63.7 &11.41&29.4\\
         Yolov9t-obb\_head & 0.907 & 0.703 &81.7&57.4&2.02&7.8 \\
         Yolov9s-obb\_head & 0.929 & 0.718 &84.4&62.0&7.36&27.6\\
         Yolov10n-obb\_head &0.916&0.707&82.7&58.4&2.67&8.2\\
         Yolov10s-obb\_head &0.933 & 0.917&84.9&63.1&9.69&28.2\\
         Yolo11n-obb &0.911 &0.702&81.8&57.7&2.65&6.6\\
         Yolov11s-obb & 0.930&0.722 &84.9 &62.3&9.70&22.3\\
    \hline
    \end{tabular}
\end{table*}

\begin{table*}[htbp]
    \caption{Annotation Type and Year Comparison}
    \label{tab:table6}
    \centering
    \footnotesize
    \setlength{\tabcolsep}{5.8pt} 
    \renewcommand{\arraystretch}{1} 
    \begin{tabular}{>{\centering\arraybackslash}m{2cm} 
                    >{\centering\arraybackslash}m{1cm} 
                    >{\centering\arraybackslash}m{9cm} 
                    >{\centering\arraybackslash}m{2cm} 
                    >{\centering\arraybackslash}m{1.5cm}}
    \hline
        Dataset & OBB & Annotation Method & Time Dimension & Year \\
    \hline
         Seaship & × & Voc2007 & × & 2018 \\
         ShipRSImageNet & \checkmark & $\theta$-based OBB: (cx, cy, w, h, $\theta$) and Point-based OBB: ($x_i$, $y_i$) i=1, 2, 3, 4 & × & 2020 \\
         HRSC2016 & \checkmark & $\theta$-based OBB: (cx, cy, w, h, $\theta$) & × & 2016 \\
         Ours & \checkmark & Point-based OBB: ($x_i$, $y_i$) i=1, 2, 3, 4 & \checkmark & 2024 \\
    \hline
    \end{tabular}
\end{table*}

\begin{table*}[htbp]
    \caption{Downstream Tasks and Technology}
    \label{tab:table7}
    \centering
    \footnotesize
    \setlength{\tabcolsep}{5.2pt} 
    \renewcommand{\arraystretch}{1.2} 
    \begin{tabular}{c >{\centering\arraybackslash}m{1.5cm} >{\centering\arraybackslash}m{2cm} 
                    >{\centering\arraybackslash}m{2cm} >{\centering\arraybackslash}m{1.5cm} 
                    >{\centering\arraybackslash}m{1.5cm} >{\centering\arraybackslash}m{2.3cm} 
                    >{\centering\arraybackslash}m{2cm}}
    \hline
        Dataset & Object Detection & Object Classification & Remote Sensing Detection & Trajectory Extraction & Trajectory Prediction & Traffic Information Analysis & Object Tracking \\
    \hline
         Seaship & \checkmark & \checkmark & & & & & \\
         ShipRSImageNet & & & \checkmark & & & & \\
         HRSC2016 & & & \checkmark & & & & \\
         Ours & \checkmark & \checkmark & & \checkmark & \checkmark & \checkmark & \checkmark \\
    \hline
    \end{tabular}
\end{table*}

\subsection{Collision Variation}
\begin{figure}
\small 
	\centering
	\includegraphics[width=0.8\linewidth]{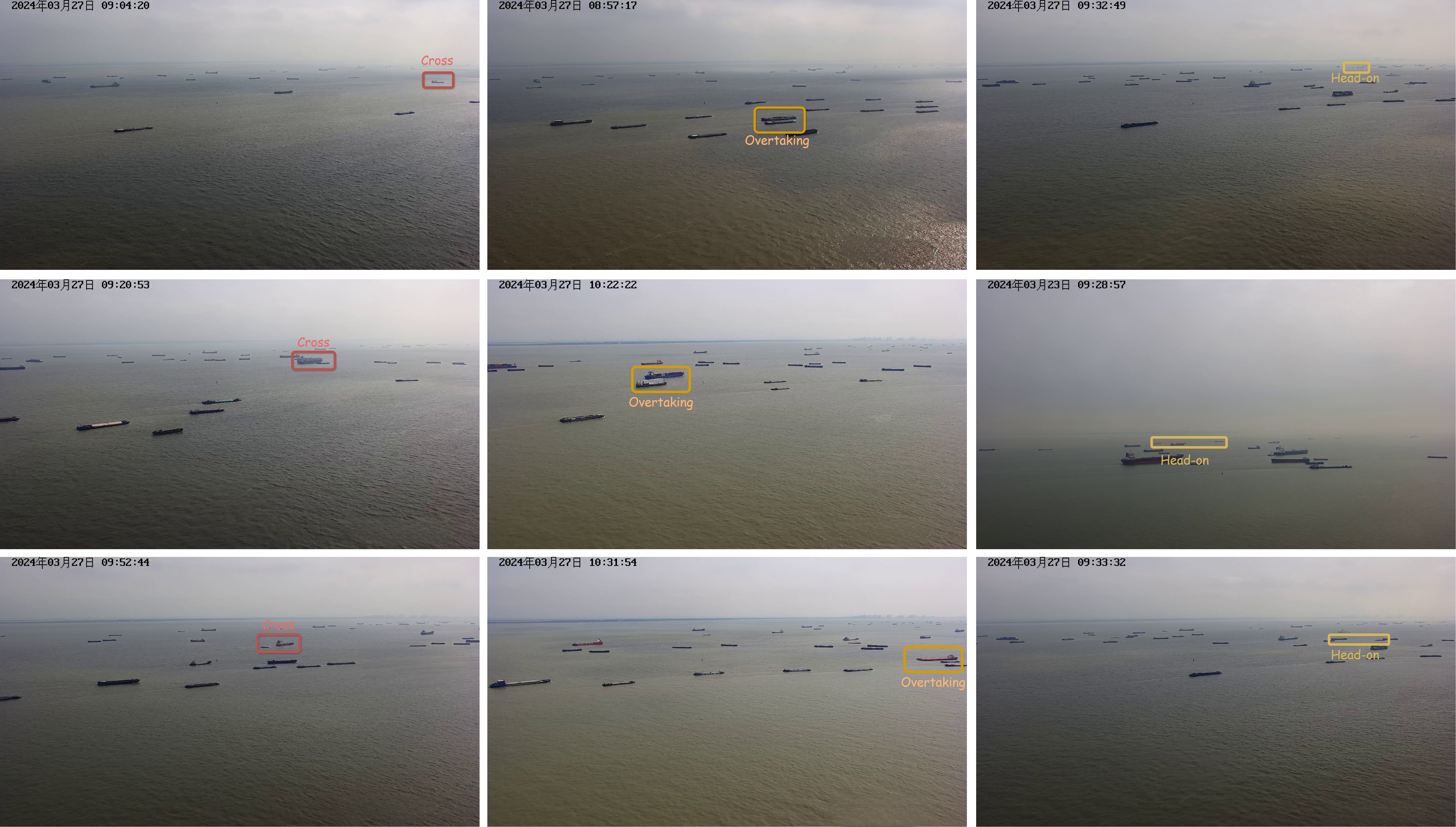}
	\caption{Examples images of different collision degrees. The first to third columns correspond to the conditions of cross, overtaking, and head-on, respectively.}
	\label{fig:fig18}
	\medskip
\end{figure}

To explore the types of collision avoidance scenarios in MID, we grouped all images with collision avoidance situations from different scenes and analyzed them. We classified the collision avoidance scenarios into three main types: crossing, overtaking, and head-on encounter. Crossing refers to two ships' paths forming a relative intersection, usually sailing on different courses that cross each other. Overtaking refers to one ship approaching and passing another from behind. Head-on encounter refers to two ships sailing directly toward each other, facing head-on. These scenarios are illustrated in Fig. \ref{fig:fig18}.

\section{Baseline Experiment and Comprehensive Analysis}
To validate the detection performance of the proposed dataset and provide reference evaluation results for related researchers, we conducted several baseline experiments. It is important to note that the experimental results for YOLOv6 \cite{li2022yolov6}, YOLOv9 \cite{wang2025yolov9}, and YOLOv10 \cite{wang2024yolov10} were obtained by replacing the detection head with an OBB detection head. Table \ref{tab:table4} presents the experimental parameter settings, while Table \ref{tab:table5} shows the experimental results.

To further illustrate the advancement and application value of the proposed dataset, we selected several representative datasets in the related field for analysis and comparison. By examining key dimensions, we demonstrate the value of our dataset for both academic research and practical tasks.
\subsection{Data Annotation and Year Comparison}
The Seaship dataset uses traditional HBB annotation, which is mainly suitable for simple scenes and single-angle detection. While the ShipRSImageNet and HRSC2016 datasets utilize OBB annotation, they are based on remote sensing images and do not capture maritime traffic information. Our dataset not only adopts OBB annotation but also incorporates a time dimension in the naming convention. Specifically, each image includes a video ID and frame ID, allowing easy tracking of the order in which targets appear, thus laying a foundation for expanding application scenarios. Table \ref{tab:table6} provides a more intuitive comparison.

\subsection{Application Scenarios and Technological Potential}
Table \ref{tab:table7} shows the applicability of different datasets for downstream tasks. Our dataset is not only suitable for coastal ship detection but also demonstrates broad technological potential across various application scenarios, including ship tracking, trajectory extraction and prediction, traffic information analysis, and autonomous vessel systems. Especially in the context of intelligent port development and autonomous vessel technology, our dataset provides a rich foundation for future research directions. In contrast, other datasets are more focused on single-task detection, with relatively limited application scenarios and technological potential.

\section{Summary and Future Outlook}
This paper presents a ship dataset based on OBB annotation, which is more flexible and accurate in handling tilted targets, complex poses, and diverse ship shapes. The dataset covers a variety of typical maritime scenarios, including ship intersection, occlusion, and small target aggregation. This diversity ensures that the model can not only handle ship detection in a single environment but also adapt to complex real-world situations, particularly excelling in ship detection in busy ports or dense maritime areas.

Additionally, to make the dataset more aligned with real-world maritime monitoring needs, our data collection includes various weather and lighting conditions, such as rainy, cloudy, foggy days, and different lighting environments. These conditions not only increase the challenge of the dataset but also enable the model to better handle ship detection tasks in diverse natural environments. The dataset particularly focuses on the annotation and detection of small target ships, which is crucial in dense ship intersections or crowded port environments where small target detection performance is critical. We provide detailed annotations for these small targets to ensure that the model has sufficient flexibility and robustness when dealing with dense or crowded scenarios.

Moreover, by using our video data and combining it with current detection algorithms, it is easy to extract ship trajectory information, enabling ship speed estimation, ship counting, and other related analyses. With these comprehensive features, our dataset demonstrates broad potential for applications in ship detection and related fields.

We plan to release version V2 of the dataset, which will expand coverage to include more diverse maritime scenarios, detailed ship annotations (such as unique IDs and navigation data), and multimodal data like AIS for enhanced ship behavior analysis. Additionally, V2 will support the development of advanced algorithms for detecting occluded and small targets in complex environments. By integrating with maritime traffic monitoring systems, this version will improve practical applications in port management and ship tracking, advancing intelligent maritime monitoring and decision-making.
\bibliographystyle{IEEEtran}
\bibliography{references}

\begin{thebibliography}{10}
\providecommand{\url}[1]{#1}
\csname url@samestyle\endcsname
\providecommand{\newblock}{\relax}
\providecommand{\bibinfo}[2]{#2}
\providecommand{\BIBentrySTDinterwordspacing}{\spaceskip=0pt\relax}
\providecommand{\BIBentryALTinterwordstretchfactor}{4}
\providecommand{\BIBentryALTinterwordspacing}{\spaceskip=\fontdimen2\font plus
\BIBentryALTinterwordstretchfactor\fontdimen3\font minus \fontdimen4\font\relax}
\providecommand{\BIBforeignlanguage}[2]{{%
\expandafter\ifx\csname l@#1\endcsname\relax
\typeout{** WARNING: IEEEtran.bst: No hyphenation pattern has been}%
\typeout{** loaded for the language `#1'. Using the pattern for}%
\typeout{** the default language instead.}%
\else
\language=\csname l@#1\endcsname
\fi
#2}}
\providecommand{\BIBdecl}{\relax}
\BIBdecl

\bibitem{martelli2021outlook}
M.~Martelli, A.~Virdis, A.~Gotta, P.~Cassar{\`a}, and M.~Di~Summa, ``An outlook on the future marine traffic management system for autonomous ships,'' \emph{IEEE Access}, vol.~9, pp. 157\,316--157\,328, 2021.

\bibitem{peters2001developments}
H.~J. Peters, ``Developments in global seatrade and container shipping markets: their effects on the port industry and private sector involvement,'' \emph{Int. J. Marit. Econ.}, vol.~3, no.~1, pp. 3--26, 2001.

\bibitem{yang2018internet}
Y.~Yang, M.~Zhong, H.~Yao, F.~Yu, X.~Fu, and O.~Postolache, ``Internet of things for smart ports: Technologies and challenges,'' \emph{IEEE Instrum. Meas. Mag.}, vol.~21, no.~1, pp. 34--43, 2018.

\bibitem{xin2023multi}
X.~Xin, Z.~Yang, K.~Liu, J.~Zhang, and X.~Wu, ``Multi-stage and multi-topology analysis of ship traffic complexity for probabilistic collision detection,'' \emph{Expert Syst. Appl.}, vol. 213, p. 118890, 2023.

\bibitem{zhou2023sidelobe}
Y.~Zhou, H.~Liu, F.~Ma, Z.~Pan, and F.~Zhang, ``A sidelobe-aware small ship detection network for synthetic aperture radar imagery,'' \emph{IEEE Trans. Geosci. Remote Sens.}, vol.~61, pp. 1--16, 2023.

\bibitem{thombre2020sensors}
S.~Thombre, Z.~Zhao, H.~Ramm-Schmidt, J.~M.~V. Garc{\'\i}a, T.~Malkam{\"a}ki, S.~Nikolskiy, T.~Hammarberg, H.~Nuortie, M.~Z.~H. Bhuiyan, S.~S{\"a}rkk{\"a} \emph{et~al.}, ``Sensors and ai techniques for situational awareness in autonomous ships: A review,'' \emph{IEEE trans. Intell. Transp. Syst.}, vol.~23, no.~1, pp. 64--83, 2020.

\bibitem{felski2020ocean}
A.~Felski and K.~Zwolak, ``The ocean-going autonomous ship—challenges and threats,'' \emph{J. Mar. Sci. Eng.}, vol.~8, no.~1, p.~41, 2020.

\bibitem{chen2024weather}
M.~Chen, J.~Sun, K.~Aida, and A.~Takefusa, ``Weather-aware object detection method for maritime surveillance systems,'' \emph{Future Gener. Comp. Sy.}, vol. 151, pp. 111--123, 2024.

\bibitem{prasad2018object}
D.~K. Prasad, C.~K. Prasath, D.~Rajan, L.~Rachmawati, E.~Rajabally, and C.~Quek, ``Object detection in a maritime environment: Performance evaluation of background subtraction methods,'' \emph{IEEE trans. Intell. Transp. Syst.}, vol.~20, no.~5, pp. 1787--1802, 2018.

\bibitem{shi2013ship}
Z.~Shi, X.~Yu, Z.~Jiang, and B.~Li, ``Ship detection in high-resolution optical imagery based on anomaly detector and local shape feature,'' \emph{IEEE Trans. Geosci. Remote Sens.}, vol.~52, no.~8, pp. 4511--4523, 2013.

\bibitem{yang2019big}
D.~Yang, L.~Wu, S.~Wang, H.~Jia, and K.~X. Li, ``How big data enriches maritime research--a critical review of automatic identification system (ais) data applications,'' \emph{Transp. Rev.}, vol.~39, no.~6, pp. 755--773, 2019.

\bibitem{barnum1986ship}
J.~Barnum, ``Ship detection with high-resolution hf skywave radar,'' \emph{IEEE J. OCEANIC ENG.}, vol.~11, no.~2, pp. 196--209, 1986.

\bibitem{robards2016conservation}
M.~Robards, G.~Silber, J.~Adams, J.~Arroyo, D.~Lorenzini, K.~Schwehr, and J.~Amos, ``Conservation science and policy applications of the marine vessel automatic identification system (ais)—a review,'' \emph{B. MAR. SCI.}, vol.~92, no.~1, pp. 75--103, 2016.

\bibitem{marino2015ship}
A.~Marino, M.~J. Sanjuan-Ferrer, I.~Hajnsek, and K.~Ouchi, ``Ship detection with spectral analysis of synthetic aperture radar: A comparison of new and well-known algorithms,'' \emph{Remote Sens.}, vol.~7, no.~5, pp. 5416--5439, 2015.

\bibitem{zhang2017ship}
Y.~Zhang, Q.-Z. Li, and F.-N. Zang, ``Ship detection for visual maritime surveillance from non-stationary platforms,'' \emph{OCEAN ENG.}, vol. 141, pp. 53--63, 2017.

\bibitem{wang2021research}
Z.~Wang, X.~Miao, Z.~Huang, and H.~Luo, ``Research of target detection and classification techniques using millimeter-wave radar and vision sensors,'' \emph{Remote Sens.}, vol.~13, no.~6, p. 1064, 2021.

\bibitem{hu2024dataset}
J.~Hu, X.~Zhi, T.~Shi, J.~Wang, Y.~Li, and X.~Sun, ``Dataset and benchmark for ship detection in complex optical remote sensing image,'' \emph{IEEE Trans. Geosci. Remote Sens.}, 2024.

\bibitem{wei2020hrsid}
S.~Wei, X.~Zeng, Q.~Qu, M.~Wang, H.~Su, and J.~Shi, ``Hrsid: A high-resolution sar images dataset for ship detection and instance segmentation,'' \emph{IEEE Access}, vol.~8, pp. 120\,234--120\,254, 2020.

\bibitem{zhang2021sar}
T.~Zhang, X.~Zhang, J.~Li, X.~Xu, B.~Wang, X.~Zhan, Y.~Xu, X.~Ke, T.~Zeng, H.~Su \emph{et~al.}, ``Sar ship detection dataset (ssdd): Official release and comprehensive data analysis,'' \emph{Remote Sens.}, vol.~13, no.~18, p. 3690, 2021.

\bibitem{su2019object}
H.~Su, S.~Wei, M.~Yan, C.~Wang, J.~Shi, and X.~Zhang, ``Object detection and instance segmentation in remote sensing imagery based on precise mask r-cnn,'' in \emph{Proc. IEEE Int. Geosci. Remote Sens. Symp.}\hskip 1em plus 0.5em minus 0.4em\relax IEEE, 2019, pp. 1454--1457.

\bibitem{su2020hq}
H.~Su, S.~Wei, S.~Liu, J.~Liang, C.~Wang, J.~Shi, and X.~Zhang, ``Hq-isnet: High-quality instance segmentation for remote sensing imagery,'' \emph{Remote Sens.}, vol.~12, no.~6, p. 989, 2020.

\bibitem{liu2017high}
Z.~Liu, L.~Yuan, L.~Weng, and Y.~Yang, ``A high resolution optical satellite image dataset for ship recognition and some new baselines,'' in \emph{Proc. Int. Conf. Pattern Recognit. Appl. Methods}, vol.~2.\hskip 1em plus 0.5em minus 0.4em\relax SciTePress, 2017, pp. 324--331.

\bibitem{zhang2021shiprsimagenet}
Z.~Zhang, L.~Zhang, Y.~Wang, P.~Feng, and R.~He, ``Shiprsimagenet: A large-scale fine-grained dataset for ship detection in high-resolution optical remote sensing images,'' \emph{EEE J. Sel. Top Appl. Earth Obs. Remote Sens.}, vol.~14, pp. 8458--8472, 2021.

\bibitem{han2021fine}
Y.~Han, X.~Yang, T.~Pu, and Z.~Peng, ``Fine-grained recognition for oriented ship against complex scenes in optical remote sensing images,'' \emph{IEEE Trans. Geosci. Remote Sens.}, vol.~60, pp. 1--18, 2021.

\bibitem{shao2018seaships}
Z.~Shao, W.~Wu, Z.~Wang, W.~Du, and C.~Li, ``Seaships: A large-scale precisely annotated dataset for ship detection,'' \emph{IEEE Trans. Multimedia.}, vol.~20, no.~10, pp. 2593--2604, 2018.

\bibitem{felzenszwalb2008discriminatively}
P.~Felzenszwalb, D.~McAllester, and D.~Ramanan, ``A discriminatively trained, multiscale, deformable part model,'' in \emph{Proc. IEEE Conf. Comput. Vision Pattern Recognit.}\hskip 1em plus 0.5em minus 0.4em\relax Ieee, 2008, pp. 1--8.

\bibitem{girshick2014rich}
R.~Girshick, J.~Donahue, T.~Darrell, and J.~Malik, ``Rich feature hierarchies for accurate object detection and semantic segmentation,'' in \emph{Proc. IEEE Conf. Comput. Vision Pattern Recognit.}, 2014, pp. 580--587.

\bibitem{girshick2015fast}
R.~Girshick, ``Fast r-cnn,'' \emph{arXiv preprint arXiv:1504.08083}, 2015.

\bibitem{ren2016faster}
S.~Ren, K.~He, R.~Girshick, and J.~Sun, ``Faster r-cnn: Towards real-time object detection with region proposal networks,'' \emph{IEEE Trans. Pattern Anal. Mach. Intell.}, vol.~39, no.~6, pp. 1137--1149, 2016.

\bibitem{redmon2016you}
J.~Redmon, ``You only look once: Unified, real-time object detection,'' in \emph{Proc. IEEE Conf. Comput. Vision Pattern Recognit.}, 2016.

\bibitem{liu2016ssd}
W.~Liu, D.~Anguelov, D.~Erhan, C.~Szegedy, S.~Reed, C.-Y. Fu, and A.~C. Berg, ``Ssd: Single shot multibox detector,'' in \emph{Proc. European Conf. Comput. Vision}.\hskip 1em plus 0.5em minus 0.4em\relax Springer, 2016, pp. 21--37.

\bibitem{ross2017focal}
T.-Y. Ross and G.~Doll{\'a}r, ``Focal loss for dense object detection,'' in \emph{Proc. IEEE Conf. Comput. Vision Pattern Recognit.}, 2017, pp. 2980--2988.

\bibitem{carion2020end}
N.~Carion, F.~Massa, G.~Synnaeve, N.~Usunier, A.~Kirillov, and S.~Zagoruyko, ``End-to-end object detection with transformers,'' in \emph{Proc. European Conf. Comput. Vision}.\hskip 1em plus 0.5em minus 0.4em\relax Springer, 2020, pp. 213--229.

\bibitem{tan2020efficientdet}
M.~Tan, R.~Pang, and Q.~V. Le, ``Efficientdet: Scalable and efficient object detection,'' in \emph{Proc. IEEE Conf. Comput. Vision Pattern Recognit.}, 2020, pp. 10\,781--10\,790.

\bibitem{li2022yolov6}
C.~Li, L.~Li, H.~Jiang, K.~Weng, Y.~Geng, L.~Li, Z.~Ke, Q.~Li, M.~Cheng, W.~Nie \emph{et~al.}, ``Yolov6: A single-stage object detection framework for industrial applications,'' \emph{arXiv preprint arXiv:2209.02976}, 2022.

\bibitem{wang2025yolov9}
C.-Y. Wang, I.-H. Yeh, and H.-Y. Mark~Liao, ``Yolov9: Learning what you want to learn using programmable gradient information,'' in \emph{Proc. European Conf. Comput. Vision}.\hskip 1em plus 0.5em minus 0.4em\relax Springer, 2025, pp. 1--21.

\bibitem{wang2024yolov10}
A.~Wang, H.~Chen, L.~Liu, K.~Chen, Z.~Lin, J.~Han, and G.~Ding, ``Yolov10: Real-time end-to-end object detection,'' \emph{arXiv preprint arXiv:2405.14458}, 2024.

\end{thebibliography}

\vfill

\end{document}